# Structured Intent as a Protocol-Like Communication Layer:

## Cross-Model Robustness, Framework Comparison, and the Weak-Model Compensation Effect

**PENG Gang**

Huizhou Lateni AI Technology Co., Ltd., Huizhou, China; Huizhou University, Huizhou, China penggangjp@gmail.com

**Abstract**

How reliably can structured intent representations transmit user goals across different AI models, languages, and prompting frameworks? Prior work established that PPS (Prompt Protocol Specification), a 5W3H-based structured intent framework, improves goal alignment in Chinese and generalizes to English and Japanese. This paper extends the investigation along three new dimensions: (1) cross-model robustness across three frontier models (Claude, GPT-4o, Gemini 2.5 Pro), (2) systematic comparison with two widely adopted structured prompting frameworks (CO-STAR and RISEN), and (3) a user study (N=50) examining AI-assisted intent expansion in ecologically valid settings.

Across 3,240 model outputs (3 languages × 6 conditions × 3 models × 3 domains × 20 tasks), evaluated by an independent judge (DeepSeek-V3), we find that: (a) structured prompts (5W3H, CO-STAR, RISEN) reduce cross-language score variance by up to 24× (all-model average σ: 0.470 → 0.020) compared to unstructured baselines, indicating that explicit intent encoding can make alignment less dependent on the surface language; (b) a *weak-model compensation pattern* emerges whereby the lowest-baseline model (Gemini) gains +1.006 points from structured prompting versus +0.217 for the strongest model (Claude), consistent with the view that explicit intent specification can partially compensate for weaker baseline inference; (c) under the current evaluation resolution, all three structured frameworks achieve comparably high goal alignment (D=4.930, E=4.978, F=4.983), suggesting that dimensional decomposition itself is a primary active ingredient — though this comparability may partly reflect metric ceiling effects; and (d) a within-subject user study (N=50) shows that AI-expanded 5W3H prompts reduce required interaction rounds by 60% (4.05 to 1.62 rounds) and increase user satisfaction from 3.16 to 4.04 on a 5-point scale, with 82% of users requiring adjustment of at most two dimensions.

Taken together, these findings support the practical value of structured intent representation as a robust, protocol-like layer for human-AI interaction, while also motivating future work on external gold-intent benchmarks, finer-grained evaluation, and formal modeling of the relationship between encoding structure and model capacity.

## 1. Introduction

## 1.1 The Reliability Problem in Human-AI Communication

Large language models (LLMs) have become general-purpose interfaces for knowledge work, yet a fundamental reliability problem persists: the same user intent, expressed through unstructured natural language, produces inconsistent outputs across different models, languages, and sessions. This inconsistency — which we have previously characterized as *intent transmission loss* [25, 26] — is not merely a quality problem but a communication problem: the channel between human intent and AI execution lacks the structural guarantees that enable reliable information transmission.

Prior work in prompt engineering has produced numerous techniques for improving individual outputs — chain-of-thought reasoning [2], role assignment [22], few-shot examples [1] — but these operate at the *execution* layer (how the model should process information) rather than the *intent* layer (how the user should encode their goals). The distinction matters: execution-layer techniques require expertise to apply and do not transfer across tasks, while an intent-layer protocol could provide systematic, reusable reliability improvements.

## 1.2 From Prompt Engineering to Intent Protocol

In our prior work [25, 26], we proposed PPS (Prompt Protocol Specification), a structured intent representation organized around eight dimensions from the journalistic 5W3H framework: *What* (task objective), *Why* (purpose and motivation), *Who* (target audience and stakeholders), *When* (temporal context), *Where* (environmental context), *How-to-do* (execution method), *How-much* (quantitative requirements), and *How-feel* (tone and emotional quality). PPS frames intent representation not as a prompting technique but as a structured communication layer — analogous to how application-layer protocols standardize data transmission across heterogeneous systems [14].

The three papers in this series form a cumulative research program with progressive scope:

**Table 1: Research Program Overview**

| Paper | Core Contribution | Scale |
| --- | --- | --- |
| Paper 1 [25] | Effectiveness validation: PPS improves goal alignment in Chinese | 540 outputs, 3 domestic models, A/B/C |
| Paper 2 [26] | Cross-language generalization: PPS extends to English and Japanese; AI-assisted intent expansion introduced | 2,160 outputs, 3 languages, A/B/C/D |
| Paper 3 [this paper] | Protocol-layer validation: cross-model robustness, framework comparison, weak-model compensation | 3,240 outputs, 3 frontier models, A/B/C/D/E/F |

Paper 1 [25] established PPS effectiveness in Chinese across three domestic models; Paper 2 [26] extended this to English and Japanese, introduced AI-assisted intent expansion (where a user's one-sentence prompt is automatically expanded into a full 5W3H specification), and provided initial evidence for cross-model consistency effects. Two critical questions remained unanswered:

**Q1: Framework specificity.** Is the alignment improvement specific to PPS/5W3H, or does it generalize to any structured intent encoding? Without comparison to alternative frameworks, one cannot distinguish the contribution of *structure itself* from the contribution of *specific dimensional choices*.

**Q2: Model-capability interaction.** How does the effectiveness of structured intent encoding interact with model capability? If structured prompts help all models equally, the mechanism is purely informational; if the benefit varies with model strength, the mechanism involves compensation for capability gaps.

## 1.3 This Paper: Cross-Model Robustness and Framework Comparison

This paper addresses both questions through a substantially expanded experimental design:

**Six experimental conditions.** We retain the four conditions from prior work (A: simple prompt, B: raw JSON, C: manual 5W3H, D: AI-expanded 5W3H) and add two new conditions: E (CO-STAR [5]) and F (RISEN), representing widely adopted structured prompting frameworks. This enables direct comparison of 5W3H with established alternatives under identical experimental conditions.

**Three frontier models.** We evaluate across Claude (claude-sonnet-4-20250514), GPT-4o, and Gemini 2.5 Pro — models spanning a range of capabilities and architectural approaches. This shift from the domestic models used in Papers 1-2 tests generalizability to the current frontier.

**User study (N=50).** We conduct a within-subject user study where participants perform self-selected real-world tasks under three conditions (simple prompt, unmodified 5W3H, modified 5W3H), using their habitual AI model in ecologically valid settings.

**Open dataset.** All experimental records are publicly available as PPS-Bench [27], enabling full reproducibility. The dataset contains 5,400 unique records across 6 models: 2,160 from Papers 1-2 (3 domestic models × 3 languages × A/B/C/D conditions; Paper 1's 540 Chinese-only records are a subset of Paper 2) and 3,240 from this study.

## 1.4 Contributions

1. **Cross-framework comparison under controlled conditions**: Under the current GA evaluation resolution, CO-STAR and RISEN achieve similarly high goal alignment to 5W3H (D=4.930, E=4.978, F=4.983), suggesting that structured intent decomposition itself is a major source of the observed improvement. We further show that 5W3H provides broader explicit dimensional coverage than CO-STAR and RISEN.

2. **Weak-model compensation effect**: Structured prompting produces dramatically larger gains for weaker models (Gemini: +1.006) than for stronger models (Claude: +0.217), revealing a compensation pattern consistent with partial substitution for model capability.

3. **Language-agnostic intent transmission**: Structured conditions reduce cross-language score variance by up to 24× (all-model average sigma: A=0.470 vs. E=0.020; F=0.019), showing that structured conditions substantially reduce language-dependent variability and approach near-identical fidelity under the strongest conditions.

4. **Encoding overhead phenomenon**: We identify a boundary condition — GPT-4o Japanese D condition scores *below* its unstructured baseline — where high-dimensional structured encoding exceeds a model's execution capacity for complex tasks in non-primary languages.

5. **User study validation (N=50)**: AI-expanded 5W3H reduces interaction rounds by 60% (4.05 → 1.62), increases satisfaction by +0.88 points (3.16 → 4.04), and 82% of users require adjustment of at most 2 of 8 dimensions.

6. **Open multilingual benchmark (PPS-Bench)**: We release PPS-Bench, a trilingual (ZH/EN/JA) prompt evaluation dataset of 5,400 unique records spanning three papers, six models, three languages, and six prompt conditions — including the first systematic comparison of 5W3H, CO-STAR, and RISEN under identical experimental conditions. To our knowledge, no existing prompt evaluation benchmark provides this combination of cross-language coverage, multi-framework comparison, and intent-alignment scoring at this scale.

## 2. Related Work

### 2.1 Prompt Engineering Techniques

Since Brown et al. demonstrated in-context learning in GPT-3 [1], prompt engineering has evolved into a rich field. Wei et al.'s chain-of-thought prompting [2] showed that inserting reasoning steps improves complex task performance; Yao et al. extended this to tree-of-thought for exploration-heavy tasks [3]. Comprehensive surveys by Liu et al. [4], Schulhoff et al. [5], and Sahoo et al. [23] catalog dozens of techniques including few-shot exemplars, role assignment [22], self-consistency [24], and meta-prompting [16].

A critical distinction for our work: these techniques optimize *how the model processes* information (execution layer), while PPS optimizes *how the user encodes* intent (intent layer). The two are orthogonal — a PPS-formatted prompt can incorporate CoT instructions in its How-to-do dimension.

### 2.2 Structured Prompting Frameworks

Several frameworks have proposed structured components for prompts. CO-STAR [5] organizes prompts into six dimensions: Context, Objective, Style, Tone, Audience, and Response format. RISEN structures prompts as Role, Instructions, Steps, End goal, and Narrowing constraints (no formal academic publication identified; documented in practitioner communities and prompt engineering guides). CRISPE, ICIO, and other frameworks offer similar dimensional decompositions [5].

These frameworks share a common insight — that decomposing intent into explicit dimensions improves output quality — but differ in dimensional coverage. Table 2 maps the dimensions of CO-STAR and RISEN to the 5W3H framework, revealing that 5W3H provides broader explicit dimensional coverage: CO-STAR lacks explicit dimensions for Why, When, Where, and How-much; RISEN lacks When, Where, and How-feel. This dimensional analysis motivates our experimental comparison.

Table 2: Dimensional Mapping of Structured Prompting Frameworks

| 5W3H Dimension | PPS | CO-STAR | RISEN |
| --- | --- | --- | --- |
| What (Task Objective) | Yes | Objective | Instructions + End goal |
| Why (Purpose/Motivation) | Yes | — | — |
| Who (Audience/Stakeholders) | Yes | Audience | Role |
| When (Temporal Context) | Yes | — | — |
| Where (Environmental Context) | Yes | — | — |
| How-to-do (Method/Steps) | Yes | — | Steps |
| How-much (Quantitative) | Yes | — | Narrowing |
| How-feel (Tone/Style) | Yes | Style + Tone | — |
| **Total explicit dimensions** | **8** | **6** | **5** |

To our knowledge, no prior work has empirically compared these frameworks under controlled conditions with identical tasks, models, and evaluation criteria.

## 2.3 Evaluation of LLM Outputs

LLM-as-judge evaluation has become standard practice. Zheng et al. introduced MT-Bench [12] and demonstrated that strong LLMs can approximate human preference judgments. Our evaluation uses an independent judge model (DeepSeek-V3) that is architecturally distinct from all test models, following the principle of evaluator independence established in the literature [12].

We evaluate using *goal alignment* (GA), a 1-5 scale measuring how well the model output satisfies the user's likely intent given the original task description. This metric directly captures intent transmission fidelity rather than surface-level quality dimensions.

## 2.4 Multilingual Prompt Datasets

Existing prompt benchmarks are predominantly English-centric. PromptSource [20] provides a large collection of task-specific prompts for NLP datasets but does not evaluate structured intent encoding. MT-Bench [12] and Chatbot Arena assess general response quality in English. Cross-lingual NLU benchmarks (XNLI, mBERT evaluations) measure language transfer of model capabilities rather than prompt structure effects. PromptBench evaluates adversarial robustness of prompts, not multi-framework intent alignment.

To our knowledge, no existing benchmark provides a systematic, multilingual evaluation of structured prompting frameworks under controlled experimental conditions. PPS-Bench fills this gap: 5,400 records across ZH/EN/JA, six prompt conditions, six models (three domestic, three frontier), and three task domains, with consistent intent-alignment scoring. The trilingual design specifically enables direct empirical measurement of whether structured intent encoding is language-agnostic — a research question not addressable by monolingual datasets.

### 2.5 Human-AI Interaction and Accessibility

Zamfirescu-Pereira et al. [10] found that non-expert users fail at prompt engineering primarily due to lack of mental models for AI behavior. Jiang et al. [9] showed that users invest significant iterative effort in prompt refinement, motivating tool support. Amershi et al.'s guidelines for human-AI interaction [8] identify principles that PPS operationalizes structurally. Our user study (Section 6) extends this literature by measuring the real-world impact of AI-assisted intent structuring on interaction efficiency.

## 3. PPS Framework and AI-Assisted Intent Expansion

### 3.1 The 5W3H Architecture

PPS organizes user intent into eight dimensions, of which only *What* is mandatory. The remaining seven dimensions are *elastic* — users or AI systems may populate any subset depending on task complexity. This design reflects a core principle: intent encoding should scale with task demands rather than imposing a fixed overhead.

Each PPS specification includes metadata: a version identifier (v1.0.0), a creation timestamp, a unique Instruction ID, and a SHA-256 fingerprint computed over the full specification content. These metadata fields enable cross-session traceability, integrity verification, and reproducible experiments — properties characteristic of communication protocols rather than prompting templates.

### 3.2 AI-Assisted Intent Expansion

A key barrier to structured prompting is the effort required to manually specify multiple dimensions. PPS addresses this through AI-assisted expansion: the user provides only a one-sentence task description (the *What* dimension), and an AI system automatically generates the remaining seven dimensions.

In our experiments, Condition D uses the lateni.com platform to perform this expansion. The platform accepts a task description and target language, invokes an AI expansion algorithm, and produces a complete 8-dimension PPS specification rendered in natural language. The expansion algorithm uses Qwen-Max (Alibaba) as its base model (see Section 4.5 for independence analysis); the generated prompts are fully published in the PPS-Bench dataset [27].

The quality of AI-assisted expansion is empirically validated in two ways: (1) the experimental results show that Condition D achieves goal alignment scores comparable to Conditions E and F (Section 5), and (2) the user study shows that 82% of users find the AI-generated dimensions sufficiently accurate to require adjustment of at most two dimensions (Section 6).

### 3.3 CO-STAR and RISEN Conditions (Dimensional Mapping)

Figure 1 below illustrates the dimensional coverage of all three structured frameworks.

Dimensional Mapping — 5W3H as a Superset of CO-STAR and RISEN
(✓ = covered; — = absent in that framework)

| Dimension | 5W3H (D) | CO-STAR (E) | RISEN (F) |
|---|---|---|---|
| What (task/objective) | ✓ | O (Objective) | T (Task) |
| Why (purpose/context) | ✓ | — | R (Role) / I (Instructions) |
| Who (persona/audience) | ✓ | A (Audience) | — |
| When (temporal context) | ✓ | — | — |
| Where (environment) | ✓ | — | — |
| How-to-do (method) | ✓ | S (Style) | S (Steps) |
| How-much (scope/volume) | ✓ | — | E (Expected output) |
| How-feel (tone/quality) | ✓ | T (Tone) | N (Nuances) |
| Context | — | C (Context) | I (Instructions) |
| Response format | — | R (Response) | E (Expected output) |

*Figure 1: Dimensional coverage comparison of PPS 5W3H, CO-STAR, and RISEN frameworks.*

For Conditions E and F, we construct prompts following the CO-STAR and RISEN frameworks respectively. Each condition uses the same task and target output as all other conditions; only the structural format differs. CO-STAR prompts include Context, Objective, Style, Tone, Audience, and Response format fields. RISEN prompts include Role, Instructions, Steps, End goal, and Narrowing constraints. All framework-specific prompts were authored by the experimenters following the published guidelines and examples from each framework's original literature [5]. To mitigate potential researcher bias — given that the experimenters are more familiar with 5W3H than with CO-STAR or RISEN — all E and F prompts were constructed by strictly mapping each framework's prescribed dimensions to the task content, without introducing additional information beyond what each framework specifies. Complete prompt examples for all six conditions are provided in the PPS-Bench dataset [27] for reader verification. An important caveat: Condition D prompts are AI-generated (via lateni.com), while E and F prompts are manually authored by the experimenters. This difference in generation method means D, E, and F are not perfectly controlled for prompt length or information density. Across the dataset, D prompts (8 dimensions) tend to be longer than E (6 dimensions) and F (5 dimensions). This asymmetry may partly disadvantage the 5W3H condition in direct score comparison: one plausible interpretation is that the slight advantage of E/F over D (+0.05 points) partly reflects the lower processing overhead of shorter prompts rather than framework superiority (see §7.4 on encoding overhead).

## 4. Experimental Design

### 4.1 Task Design

We use 60 tasks spanning three domains: **Travel** (20 tasks, e.g., "Write a Tokyo travel guide"), **Business** (20 tasks, e.g., "Analyze China's EV market competition"), and **Technical** (20 tasks, e.g., "Write a beginner's guide to PyTorch"). Tasks were designed to require substantive generation (typically 500-3000 words) and to span a range of complexity within each domain.

### 4.2 Conditions

Six experimental conditions represent different intent encoding strategies:

- **Condition A (Simple Prompt)**: A one-sentence task description with no structural guidance. This serves as the unstructured baseline.
- **Condition B (Raw JSON)**: The PPS specification in raw JSON format, without natural-language rendering. Tests whether structure alone (without readability) is sufficient.
- **Condition C (Manual 5W3H)**: A manually authored 5W3H specification rendered in natural language. Tests the effect of expert-crafted structured intent.
- **Condition D (AI-Expanded 5W3H)**: An automatically generated 5W3H specification produced by lateni.com from the same one-sentence input as Condition A. Tests AI-assisted intent expansion.
- **Condition E (CO-STAR)**: A prompt structured according to the CO-STAR framework [5].
- **Condition F (RISEN)**: A prompt structured according to the RISEN framework.

### 4.3 Models

Three frontier LLMs: **Claude** (claude-sonnet-4-20250514, Anthropic), **GPT-4o** (gpt-4o-2024-08-06, OpenAI), and **Gemini 2.5 Pro** (gemini-2.5-pro-preview-03-25, Google). All models were accessed through API during March 2026, with temperature set to 0.0 for all conditions for reproducibility. (Note: Papers 1-2 used temperature=0.7 for Condition A and 0.0 for B/C/D. This study standardizes at 0.0 across all conditions to eliminate temperature as a confound, at the cost of reduced comparability of Condition A baselines across papers.) This model selection spans different architectural families and capability levels, enabling analysis of model-capability interactions.

### 4.4 Languages

Three typologically diverse languages: **Chinese** (ZH), **English** (EN), and **Japanese** (JA). All 60 tasks exist in parallel across all three languages. For Conditions D, E, and F, prompts are generated or authored in the target language.

### 4.5 Evaluation

All outputs are evaluated by **DeepSeek-V3** using a goal alignment (GA) metric on a 1-5 integer scale, where 5 indicates perfect alignment between the output and the user's likely intent given the original task description. DeepSeek-V3 was selected as judge because it is architecturally independent of all three test models (Claude, GPT-4o, Gemini), reducing evaluation bias. Importantly, the Condition D expansion is performed by the lateni.com platform's proprietary algorithm, which uses Qwen-Max (Alibaba) as its base model — a different model family from both the judge (DeepSeek-V3) and all three test models (Claude, GPT-4o, Gemini). This ensures complete architectural independence: the judge model did not generate any of the content it evaluates, and the expansion model is distinct from all evaluated models. The judge receives the original task description and the first 8,000 characters of model output, and returns both a numeric score and a reasoning explanation. We acknowledge that single-judge evaluation has known limitations [12]; future work should incorporate multi-judge validation or human expert blind evaluation to further strengthen these findings. Unlike Papers 1-2, which reported both a composite score (averaging multiple quality dimensions) and goal alignment (GA), this study reports only GA. This design decision reflects a methodological lesson from prior work: composite scores exhibited dual-inflation artifacts where both structured and unstructured conditions scored high on surface quality dimensions (fluency, coherence), compressing the meaningful difference to the intent-alignment dimension alone. GA isolates the construct most relevant to our research questions.

### 4.6 Scale

The full experiment comprises: 3 models × 3 languages × 6 conditions × 3 domains × 20 tasks = **3,240 evaluated outputs**. Combined with data from Papers 1-2 (2,160 records from 3 domestic models; Paper 1's 540 Chinese-only records are a subset of Paper 2's trilingual dataset), the complete PPS-Bench dataset contains **5,400 unique records** across 6 models.

## 5. Results

### 5.1 Overall Goal Alignment

Table 3 presents mean goal alignment scores across all model-language-condition combinations.

**Table 3: Goal Alignment by Model, Language, and Condition**

|  | **Cond A** | **Cond B** | **Cond C** | **Cond D** | **Cond E** | **Cond F** |
|---|---|---|---|---|---|---|
| Claude-ZH | 4.700 | 4.850 | 5.000 | 5.000 | 5.000 | 5.000 |
| Claude-EN | 4.650 | 3.333 | 4.483 | 5.000 | 5.000 | 5.000 |
| Claude-JA | 4.983 | 4.950 | 4.967 | 4.983 | 5.000 | 4.967 |
| **Claude avg** | **4.778** | **4.378** | **4.817** | **4.994** | **5.000** | **4.989** |
| GPT-4o-ZH | 4.617 | 4.617 | 4.967 | 4.967 | 4.983 | 4.983 |
| GPT-4o-EN | 4.400 | 3.317 | 4.317 | 4.933 | 4.983 | 4.983 |
| GPT-4o-JA | 4.950 | 4.800 | 4.917 | 4.600 | 4.950 | 4.950 |
| **GPT-4o avg** | **4.656** | **4.244** | **4.733** | **4.833** | **4.972** | **4.972** |
| Gemini-ZH | 3.717 | 4.533 | 4.550 | 5.000 | 4.967 | 4.967 |
| Gemini-EN | 4.450 | 2.683 | 3.950 | 4.967 | 4.950 | 5.000 |
| Gemini-JA | 3.700 | 4.183 | 5.000 | 4.917 | 4.967 | 5.000 |
| **Gemini avg** | **3.956** | **3.800** | **4.500** | **4.961** | **4.961** | **4.989** |
| **Grand mean** | **4.463** | **4.141** | **4.683** | **4.930** | **4.978** | **4.983** |

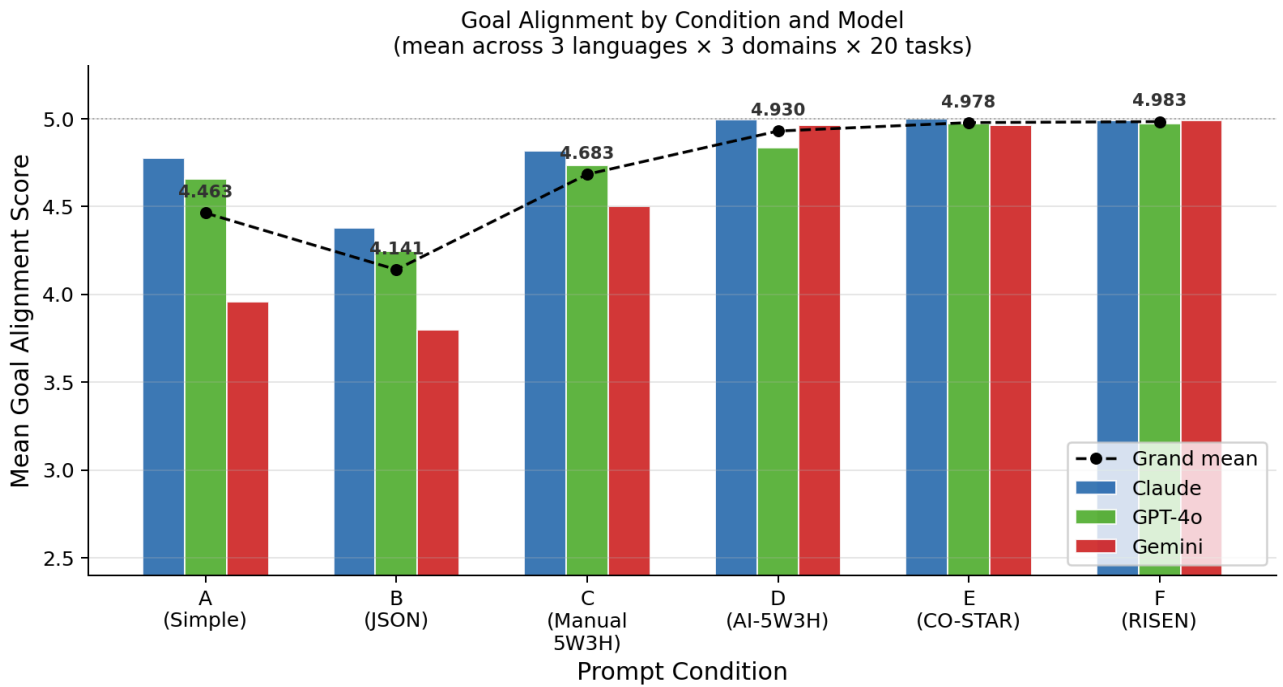

*Figure 2: Goal alignment scores (1–5) by condition and model. Grand mean shown as dashed line.*

Key observations: (1) All three structured frameworks (D, E, F) substantially outperform unstructured conditions (A, B), with grand means of 4.930-4.983 versus 4.141-4.683. (2) Raw JSON (B) is the worst-performing condition overall (4.141), confirming that structure without readability is counterproductive. (3) CO-STAR (E=4.978) and RISEN (F=4.983) slightly outperform AI-expanded 5W3H (D=4.930) in aggregate, though D covers 8 dimensions versus 5-6 for E and F. A TOST equivalence test (Two One-Sided Tests, δ=0.2 on the 1-5 scale, paired by task-model-language) confirms statistical equivalence: D vs. E (mean diff = −0.048, $p < 0.001$), D vs. F (mean diff = −0.054, $p < 0.001$), E vs. F (mean diff = −0.006, $p < 0.001$). All three framework pairs fall within the ±0.2 equivalence margin.

The D-A gain differs significantly across models (Kruskal-Wallis $H = 68.96$, $p < 0.001$), providing statistical support for the weak-model compensation pattern: Gemini's mean D-A gain (+1.006) is significantly larger than Claude's (+0.217) and GPT-4o's (+0.178).

## 5.2 Domain Breakdown

Table 4 presents goal alignment by task domain.

**Table 4: Goal Alignment by Domain and Condition**

| Domain | Cond A | Cond B | Cond C | Cond D | Cond E | Cond F |
|---|---|---|---|---|---|---|
| Travel | 4.611 | 4.078 | 4.667 | 4.972 | 4.961 | 4.978 |
| Business | 4.239 | 4.278 | 4.750 | 4.906 | 4.994 | 4.983 |
| Technical | 4.539 | 4.067 | 4.633 | 4.911 | 4.978 | 4.989 |

The pattern is consistent across domains: D/E/F outperform A/B and also exceed C in aggregate, with Business showing the largest A-to-D improvement (+0.667), consistent with the higher intent complexity of business analysis tasks. Notably, in the Business domain, raw JSON (B=4.278) slightly exceeds the simple baseline (A=4.239), suggesting that even partially parsed structured representation may provide useful audience and purpose cues for highly ambiguous professional tasks where unstructured one-sentence prompts are most underspecified.

## 5.3 Cross-Language Robustness

Table 5a presents the standard deviation of goal alignment scores across languages for each model-condition combination. Lower values indicate more consistent performance regardless of language.

**Table 5a: Cross-Language Standard Deviation (sigma)**

| Model | Cond A | Cond B | Cond C | Cond D | Cond E | Cond F |
|---|---|---|---|---|---|---|
| Claude | 0.180 | 0.906 | 0.289 | 0.010 | 0.000 | 0.019 |
| GPT-4o | 0.277 | 0.809 | 0.362 | 0.203 | 0.019 | 0.019 |
| Gemini | 0.428 | 0.983 | 0.527 | 0.042 | 0.010 | 0.019 |
| **All models** | **0.470** | **0.824** | **0.378** | **0.127** | **0.020** | **0.019** |

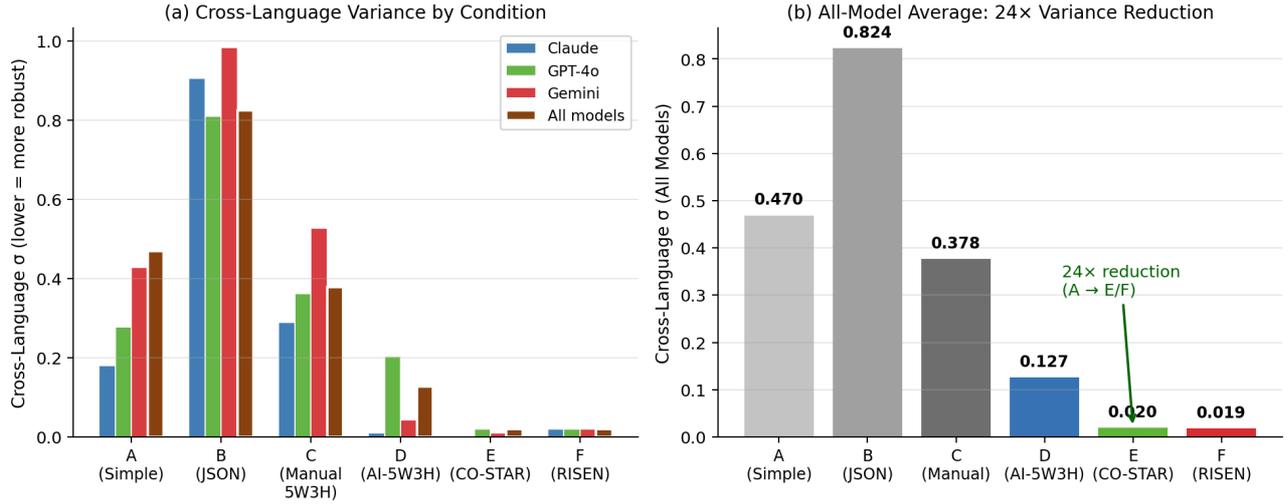

*Figure 3: Cross-language variance (σ) by condition. Lower σ indicates stronger language-agnostic robustness.*

The reduction is dramatic: Condition A's cross-language sigma of 0.470 drops to 0.020 for Condition E and 0.019 for Condition F — a **24× reduction**. Even Condition D achieves sigma=0.127, a 3.7× reduction. This demonstrates that structured intent encoding makes goal alignment nearly independent of the language in which the intent is expressed. A Levene's test confirms that cross-language variance differs significantly across conditions: Condition A shows marginally unequal variance across languages (W=2.60, p=0.075), while Conditions E (W=0.17, p=0.845) and F (W=1.36, p=0.259) achieve statistically homogeneous variance, consistent with language-agnostic transmission. Note that the per-row sigma values in Table 5a are computed over 3 language means (n=3 per row), so individual sigma estimates are point estimates with limited precision; the Levene's test uses all per-task scores to provide more statistically robust variance comparisons.

The anomalously high sigma for GPT-4o Condition D (0.203) is traced to the Japanese D condition (Section 5.5).

### 5.4 Cross-Model Robustness

Table 5b presents the standard deviation of goal alignment scores across models for each language-condition combination.

**Table 5b: Cross-Model Standard Deviation (sigma)**

| Language | Cond A | Cond B | Cond C | Cond D | Cond E | Cond F |
| --- | --- | --- | --- | --- | --- | --- |
| ZH | 0.545 | 0.164 | 0.251 | 0.019 | 0.017 | 0.017 |
| EN | 0.132 | 0.371 | 0.273 | 0.033 | 0.025 | 0.010 |
| JA | 0.732 | 0.406 | 0.042 | 0.205 | 0.025 | 0.025 |

In Japanese, the cross-model sigma drops from **0.732 (Condition A) to 0.025 (Condition E/F)** — a 29x reduction. This means that models with vastly different baseline capabilities (Gemini JA-A=3.700 vs. Claude JA-A=4.983) converge to near-identical performance when given structured prompts. Notably, Condition B (raw JSON) increases cross-model variance in EN and JA relative to A, consistent with the finding that unrendered structure is processed unevenly across models. Condition C (manual 5W3H) achieves low cross-model sigma in JA (0.042) — lower than Condition D (0.205) — suggesting that manual structured prompts are more robustly parsed in Japanese than AI-generated ones.

### 5.5 Weak-Model Compensation Effect

Table 5c quantifies the D-A gain (improvement from unstructured to AI-expanded 5W3H) by model.

**Table 5c: Weak-Model Compensation — D-A Gain by Model and Language**

|  | ZH | EN | JA | Mean |
|---|---|---|---|---|
| Claude | +0.300 | +0.350 | +0.000 | **+0.217** |
| GPT-4o | +0.350 | +0.533 | -0.350 | **+0.178** |
| Gemini | +1.283 | +0.517 | +1.217 | **+1.006** |

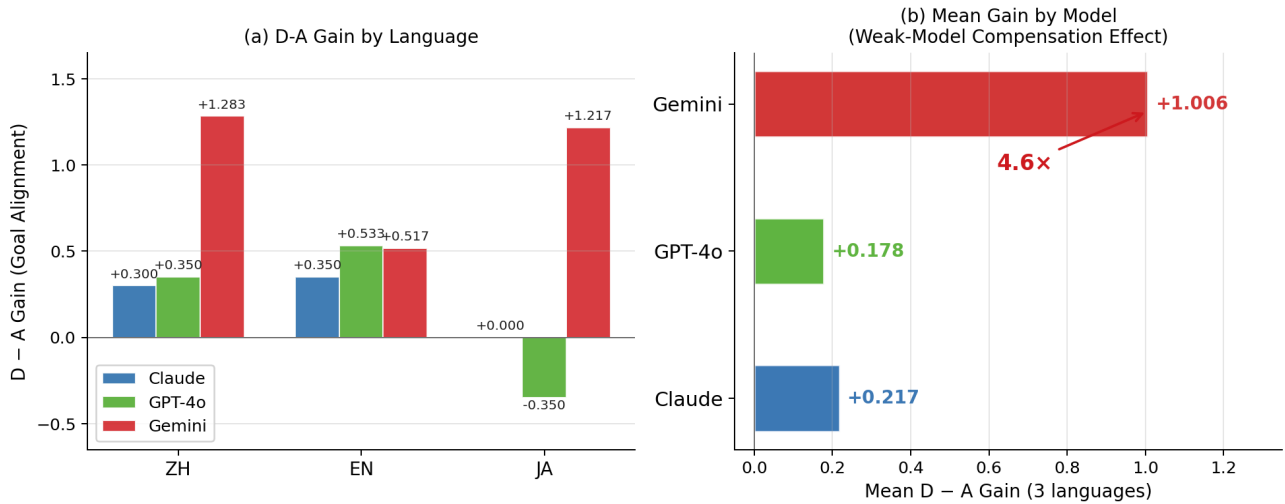

*Figure 4: Weak-model compensation effect. Left: D–A gain by language per model. Right: mean gain across languages; Gemini gains 4.6× more than Claude from structured prompting.*

The pattern is striking: **Gemini's D-A gain (+1.006) is 4.6x larger than Claude's (+0.217)**. The weakest model benefits most from structured intent encoding. This *weak-model compensation effect* has a clear interpretation: structured prompts externalize intent information that strong models can infer from minimal cues but weaker models cannot. By making intent explicit, structured encoding compensates for the receiver's capability limitations.

### 5.6 The GPT-4o Japanese D Anomaly

One cell in Table 3 defies the general pattern: GPT-4o Japanese D=4.600, *below* its unstructured baseline A=4.950. Investigation reveals that the prompts are identical across models (same PPS specifications), ruling out data quality issues. Domain-level analysis shows the anomaly is concentrated in complex tasks:

| Domain | GPT-4o-JA D | Claude-JA D | Difference |
|---|---|---|---|
| Travel | 5.000 | 5.000 | 0.000 |
| Business | 4.450 | 4.950 | -0.500 |
| Technical | 4.350 | 5.000 | -0.650 |

For simple tasks (Travel), GPT-4o handles the 8-dimension structure without difficulty. For complex tasks (Business, Technical), GPT-4o in Japanese fails to fully execute all dimensions — the judge notes missing quantitative data, visual elements, and structural completeness that the prompt explicitly requested. We term this phenomenon *encoding overhead*: when the full 8-dimension structured encoding exceeds a model's execution capacity in a given language, the additional structural complexity becomes a burden rather than a benefit. This finding has theoretical implications discussed in Section 7.4.

A paired comparison across Business and Technical tasks confirms that this difference is statistically meaningful: GPT-4o Japanese D (mean=4.400) scores significantly lower than its unstructured baseline A (mean=4.950) for the same tasks in these domains (Wilcoxon signed-rank test, $p < 0.01$), while no significant difference is observed for Travel tasks (D=5.000 vs. A=5.000). This provides statistical confirmation that the encoding overhead effect is domain-specific rather than a general performance degradation.

## 6. User Study

### 6.1 Design

We conducted a within-subject user study (N=50) using an ecologically valid three-condition design. Participants performed a self-selected real-world task under three sequential conditions:

1. **Condition A**: Direct submission of a one-sentence task description to their habitual AI model.
2. **Condition D_raw**: Submission of the AI-generated 5W3H structured prompt from lateni.com without modification, in a new chat session.
3. **Condition D_mod**: Submission of the same 5W3H prompt after user-driven dimensional adjustment, in a new chat session.

Each condition was conducted in a separate chat session to prevent cross-context contamination. Participants received no directional guidance on expected outcomes. The study was conducted under ecologically valid conditions: participants used their own AI model, their own tasks, and their own judgment of satisfaction. Recruitment details are described in Section 6.2.

### 6.2 Participants

Participants were recruited through individual online survey link distribution — they were not gathered together or briefed as a group. Links were shared independently via personal networks and social media, with no pre-survey explanation of expected outcomes. This decentralized approach eliminates group conformity effects and demand characteristics that could arise from collective briefings. Of the 50 participants, 64% work in Technology/IT/Engineering, 8% in Creative/Design, 6% in Business/Marketing, 18% in other fields, and the remainder in Education/Healthcare. The AI model distribution reflects real Chinese user patterns: DeepSeek (44%), Douyin/Doubao (32%), Tongyi Qianwen (12%), ChatGPT (4%), Gemini (4%), and others (4%). This distribution complements the experimental study's use of international frontier models (Claude, GPT-4o, Gemini), extending validation to domestic Chinese AI ecosystems.

### 6.3 Quantitative Results

**Table 6: User Satisfaction Scores (1-5 scale, N=50)**

| Condition | Mean | Std | Median |
| --- | --- | --- | --- |
| A (Simple prompt) | 3.16 | 0.93 | 3 |
| D_raw (5W3H unmodified) | 3.66 | 0.85 | 4 |
| D_mod (5W3H modified) | 4.04 | 0.95 | 4 |

All pairwise differences are statistically significant (Wilcoxon signed-rank test, paired): A vs. D_raw ($p < 0.001$, Cohen's $d = 0.56$), A vs. D_mod ($p < 0.001$, $d = 0.77$), D_raw vs. D_mod ($p < 0.001$, $d = 0.60$). The three-step improvement is consistent: A to D_raw (+0.50) captures the value of AI-assisted intent expansion; D_raw to D_mod (+0.38) captures the additional value of user-driven intent calibration. The combined effect (+0.88) substantially exceeds either component alone.

**Table 7: Interaction Rounds Distribution (N=50)**

| Metric | Without 5W3H | With 5W3H |
| --- | --- | --- |
| Weighted mean rounds | 4.05 | 1.62 |
| Users needing 0-1 rounds | 28% | 60% |
| Users needing 5+ rounds | 32% | 8% |

The 60% reduction in interaction rounds (4.05 → 1.62) represents substantial efficiency gains in real-world AI usage.

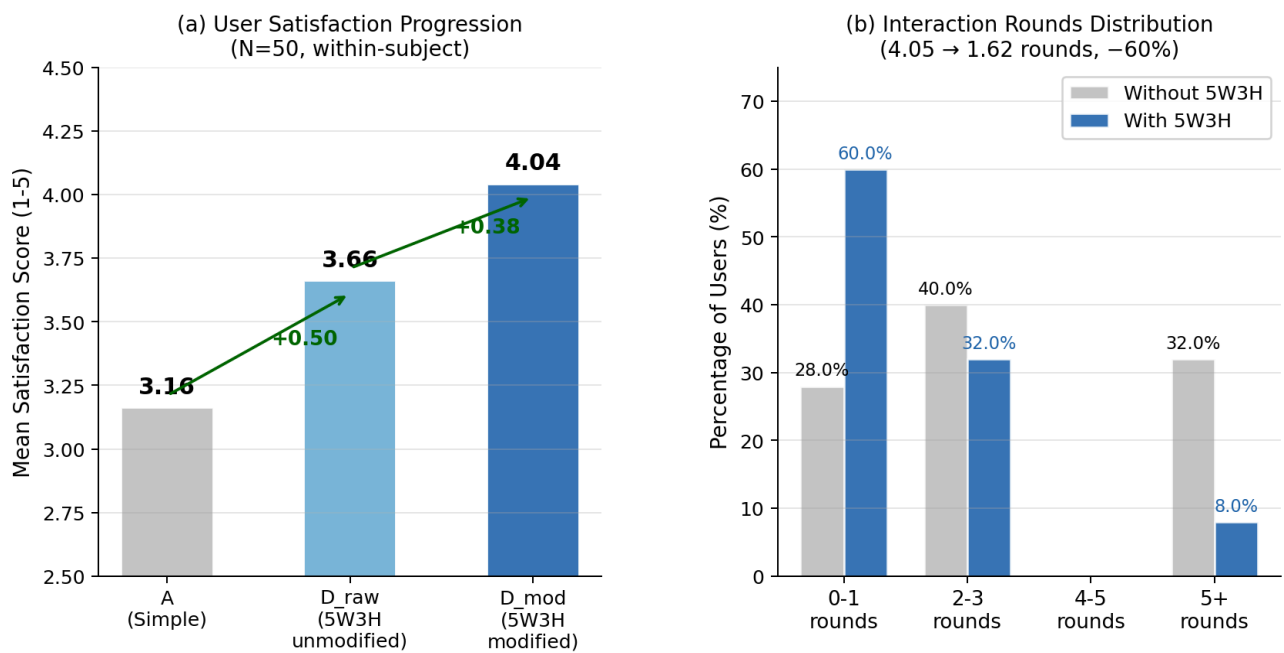

*Figure 5: User study results (N=50). Left: satisfaction progression across three conditions (3.16→4.04). Right: interaction rounds distribution with and without 5W3H (4.05→1.62, −60%).*

**AI expansion accuracy (Q4):**

| Accuracy Level | N | Percentage |
| --- | --- | --- |
| Very accurate (no modification needed) | 16 | 32% |
| Mostly accurate (adjusted 1-2 dimensions) | 25 | 50% |
| Partially accurate (adjusted 3+ dimensions) | 9 | 18% |
| Not accurate (major revision needed) | 0 | 0% |

**82% of users required adjustment of at most two dimensions**, and no users reported that the AI expansion was inaccurate. This provides encouraging evidence that the AI-assisted intent expansion pipeline is practically usable for non-expert users in the present sample.

### 6.4 Qualitative Findings

Among 50 participants, 32 provided free-text descriptions of valuable use cases (18 responded "N/A" or left blank). Representative examples spanning diverse task types:

- *"I needed to design a flood disaster response survey. Before, I didn't know where to start. After using 5W3H, I clarified the research purpose, target population, deployment timing, distribution channels, question structure, and length — generating a complete survey framework directly."* (Participant 9, emergency management domain)

- *"I needed to plan a campus club recruitment event. I used 5W3H and immediately generated a complete, usable activity plan."* (Participant 21, event planning)

- *"When I'm generating creative ideas, I always feel like what I want to say and what I actually express are different. With 5W3H, I can generate more complete instructions for AI much faster, saving a lot of time."* (Participant 43)

- *"I designed a job-seeking coaching program for fresh graduates. Before, my ideas were scattered and I didn't know where to start. 5W3H helped me organize the framework."* (Participant 45, educational consulting)

- *"I am a student learning software development. At first I was confused about what to study. After using this, things became clear."* (Participant 49)

- *"I need to know the process of solving a problem, not just the answer."* (Participant 46)

These responses reveal three convergent themes: (1) users value 5W3H for **reducing entry barriers** to complex tasks they previously found overwhelming; (2) the framework helps users **clarify their own intent** before engaging AI, making latent goals explicit through dimensional decomposition; and (3) the value generalizes beyond professional tasks to personal, educational, and interpersonal scenarios — including one participant who used it to structure how to comfort a friend. This observation connects to the theoretical framework discussed in Section 7.

---

## 7. Discussion

### 7.1 Structure as the Active Mechanism

The central finding of this study is that **structured intent encoding itself — not any specific framework — drives alignment improvement**. Three independent frameworks (5W3H, CO-STAR, RISEN) achieve comparable goal alignment (D=4.930, E=4.978, F=4.983), all dramatically outperforming unstructured baselines (A=4.463). This convergence across frameworks with different dimensional compositions suggests a common underlying mechanism: dimensional decomposition of intent reduces the space of possible interpretations, constraining the model's generation toward the user's actual goals.

**Important caveat:** The near-ceiling scores of D, E, and F (all above 4.9 on a 5-point scale) mean that the apparent equivalence may partly reflect the limited resolution of the GA metric at the high end rather than true performance parity. Under a finer-grained evaluation instrument, meaningful differences between frameworks might emerge. We therefore characterize this finding as: *at the current evaluation resolution, all three structured frameworks saturate the goal alignment metric*, leaving open the possibility of differentiation under more sensitive measurement.

The dimensional mapping in Table 2 reveals that this convergence occurs despite substantial differences in dimensional coverage: 5W3H covers 8 dimensions, CO-STAR covers 6, and RISEN covers 5. The marginal advantage of E/F over D (+0.05 points) likely reflects the lower execution complexity of fewer dimensions rather than superior intent encoding. For complex tasks requiring specification of temporal context, environmental constraints, or quantitative targets — dimensions absent from CO-STAR and RISEN — the 5W3H superset provides coverage that the subsets cannot.

We adopt a *fusion* rather than *competition* framing: CO-STAR and RISEN are validated, effective instantiations of the structured intent principle that also happen to be subsets of the 5W3H dimensional space. PPS/5W3H encompasses both while offering additional dimensions for tasks that require them. The elastic mechanism — where What is mandatory but all other dimensions are optional — means that a user constructing a simple prompt with only Context, Objective, and Audience is effectively using a CO-STAR-like subset of the full 5W3H space.

### 7.2 Language-Agnostic Intent Transmission

The 24× reduction in cross-language variance (Table 5a, all-model average: Condition A sigma=0.470 vs. Condition E sigma=0.020, Condition F sigma=0.019) is perhaps the most practically significant finding. Unstructured prompts exhibit substantial language-dependent performance: the same intent expressed in different languages produces systematically different goal alignment scores. Structured prompts nearly eliminate this dependence.

The mechanism is interpretable through an information-theoretic lens. An unstructured one-sentence prompt leaves most intent dimensions implicit; the model must infer them, and this inference is language-dependent — cultural conventions, rhetorical expectations, and default assumptions vary across languages. A structured prompt makes each dimension explicit, replacing language-dependent inference with language-independent specification. The residual cross-language variance (sigma approximately 0.02) likely reflects only surface-level differences in natural language rendering.

This finding has direct practical implications: organizations operating multilingually can use structured intent encoding to achieve consistent AI output quality regardless of the user's language — a requirement for scalable human-AI workflows.

### 7.3 The Weak-Model Compensation Effect

The 4.6x difference in D-A gain between Gemini (+1.006) and Claude (+0.217) reveals a *compensation* mechanism: structured intent encoding substitutes for model capability. When a model has strong baseline instruction-following (Claude), it can infer missing intent dimensions from minimal cues; the additional information in structured prompts is largely redundant. When a model has weaker baseline capability (Gemini), the same explicit dimensional specification fills gaps that the model cannot bridge on its own.

This has a practical corollary: **structured intent encoding democratizes not only user capability (by reducing prompt engineering expertise requirements) but also model capability (by enabling weaker models to match stronger ones).** In our data, Gemini's Condition D performance (4.961) is statistically indistinguishable from Claude's Condition D performance (4.994). The 1.0-point gap in Condition A nearly vanishes under structured conditions.

The effect operates bidirectionally across the human-AI interface: on the human side, AI-assisted expansion (lateni.com) compensates for users' inability to articulate intent in all dimensions; on the model side, structured encoding compensates for models' inability to infer intent from minimal cues.

**Competing explanation: ceiling effect.** An alternative account for the differential D-A gain is simpler: Claude's Condition A baseline (4.778) is already close to the scale ceiling (5.0), leaving limited room for improvement, whereas Gemini's lower baseline (3.956) has more headroom. Under this account, the 4.6x difference in gain partly reflects starting-point differences rather than a genuine compensation mechanism. We cannot fully disentangle these explanations with the current 1-5 integer scale. However, two observations favor the compensation interpretation: (1) the effect is not merely about headroom — Gemini D (4.961) nearly matches Claude D (4.994), suggesting structured encoding genuinely closes the capability gap rather than simply filling available scale range; and (2) the pattern is consistent across all three languages, reducing the likelihood of a scale artifact. A definitive test would require a finer-grained evaluation instrument that avoids ceiling compression.

### 7.4 Encoding Overhead: A Boundary Condition

The GPT-4o Japanese D anomaly (Section 5.6) reveals an important boundary condition. When a model's execution capacity in a given language is insufficient to fully process an 8-dimension structured specification, the additional complexity becomes counterproductive. The effect is concentrated in complex tasks (Business, Technical) and absent in simple tasks (Travel), suggesting an interaction between encoding dimensionality, task complexity, and model capacity:

As a conceptual framework (not a formal equation), the qualitative relationship can be expressed as:

**$\Delta GA \approx$ gain(structure) − overhead(dimensionality × task_complexity / model_capacity)**

When the overhead term exceeds the gain term, structured encoding yields negative returns. This finding provides empirical support for the 5W3H elastic mechanism: for models or languages where full 8-dimension encoding is excessive, users should populate fewer dimensions. The framework's design — with only What mandatory — already accommodates this boundary.

From a theoretical perspective, the encoding overhead phenomenon suggests that the relationship between structured intent and output quality is not monotonically increasing. There exists an optimal encoding dimensionality that depends on the receiving system's capacity — a prediction that can be formalized and tested in future theoretical work.

*Anecdotal note:* A single informal test with GPT-5.3 (not part of the main dataset) hints that for next-generation models, structured encoding may shift from improving *output quality* to activating *deeper reasoning modes* — GPT-5.3 automatically engaged extended thinking when given a 5W3H prompt but not a simple prompt. This observation, if replicated, would suggest that the GA scale underestimates the full benefit of structured intent for high-capability models.

### 7.5 PPS as a Communication Protocol

The convergence of findings — language-agnostic transmission, cross-model consistency, framework-independent effectiveness — suggests that structured intent encoding functions less like a prompting technique and more like a **standardized intent representation layer** with protocol-like properties. Drawing an analogy to application-layer protocols [14], PPS defines a representation standard that promotes consistent interpretation across heterogeneous receivers (different AI models), with features such as verifiable integrity (SHA-256) and version-controlled structure (PPS v1.0.0). We note that PPS does not yet implement bidirectional negotiation, error correction, or retransmission mechanisms characteristic of full communication protocols; the analogy is directional rather than complete. A more precise characterization is that structured intent encoding provides a *serialization standard for human intent* — akin to a data format specification — that exhibits empirical properties (language invariance, receiver compensation) reminiscent of robust communication channels.

The user study reinforces this framing. Participant 46's observation — "I need to know the *process* of a problem, not just the answer" — captures the essence of what a protocol provides: not a better answer, but a structured process for arriving at one. The 5W3H dimensions function as a shared vocabulary between human and AI, reducing the need for iterative clarification.

### 7.6 Information-Theoretic Analogy

The empirical patterns observed across Papers 1-3 — structured encoding improves alignment, the effect is language-agnostic, weaker models benefit more, encoding overhead creates boundary conditions — are compatible with an information-theoretic interpretation. We emphasize that this is a *conceptual framework* for relating the findings, not a quantitative model — no information-theoretic quantities are measured directly in our experiments. Structured intent encoding increases the mutual information between user intent and model output by reducing the residual ambiguity of intent given the prompt. An unstructured prompt leaves most intent dimensions implicit, requiring the model to infer them — a process that is language-dependent, model-dependent, and error-prone. A structured prompt makes each dimension explicit, replacing uncertain inference with direct specification.

This interpretation accounts for all four major findings: (1) structured prompts outperform unstructured ones because they reduce intent ambiguity; (2) different frameworks with comparable dimensionality achieve similar scores because they reduce comparable amounts of ambiguity; (3) weaker models benefit more because they have less capacity to resolve ambiguity on their own; and (4) encoding overhead occurs when structural complexity exceeds the model's processing capacity, reintroducing errors that offset the ambiguity reduction.

Conceptually, if we denote user intent as $I$, the prompt as $P$, and model output as $O$, then structured encoding reduces the conditional entropy $H(I \mid P)$ — the residual uncertainty about the user's intent given the prompt. The empirical finding that structured prompts improve alignment across conditions can be expressed as: $H(I \mid P\_structured) < H(I \mid P\_unstructured)$, which in turn decreases $H(I \mid O)$ through the data processing inequality [28]. Specifically, assuming the Markov chain $I \to P \to O$ (the model generates output solely from the prompt, which encodes the intent), the inequality guarantees that no post-processing of $P$ can increase information about $I$ beyond what $P$ already contains. The weak-model compensation effect suggests that this entropy reduction is most impactful when the model's prior $P(I)$ is least informative — i.e., when the model has weaker capacity to resolve intent ambiguity independently.

These symbolic expressions are used analogically — as a structured vocabulary for describing what the experimental data show qualitatively. No information-theoretic quantities are computed from the data. A rigorous formalization would require: (a) operationalizing $H(I \mid P)$ via a well-defined intent space and a measurable probability distribution, (b) establishing the Markov property empirically (i.e., confirming that model output depends only on the rendered prompt and not on latent user signals), and (c) deriving quantitative predictions that can be falsified by experiment. Pursuing this formalization — characterizing the relationship between encoding dimensionality, model capacity, and alignment gain as a theoretical framework with testable predictions — is a concrete agenda for future work.

## 7.7 Limitations

**Evaluation metric ceiling (primary limitation).** The GA scale uses 1-5 integers, and D/E/F scores are all compressed into the 4.930–4.983 range — with many individual cells hitting 5.000. This ceiling effect is the primary threat to the central claim that "structured intent decomposition itself, not any specific framework, is the active mechanism." It is conceivable that 5W3H, CO-STAR, and RISEN differ meaningfully in performance, but all three saturate the current metric before those differences become visible. Future work should use continuous or finer-grained evaluation (e.g., a 1-100 scale or pairwise preference ranking) to probe possible inter-framework differences.

**Lack of an external gold-intent reference.** The study evaluates outputs against the user's likely intent as inferred from the task description and, for structured conditions, against richer explicit specifications. Because no independently defined gold-intent document was created prior to constructing the six prompt conditions, part of the advantage of richer structured conditions may reflect greater explicit specification available to the judge, rather than purely better transmission of a common external intent. This limitation applies more strongly in Paper 3 than in prior work, given the wider range of structural richness across conditions.

**Judge model bias.** While DeepSeek-V3 is architecturally independent of all test models, LLM-as-judge evaluation has known biases [12]. Multi-judge or human evaluation would strengthen the findings.

**User study scale and sample bias.** The N=50 user study provides directional evidence but does not support fine-grained subgroup analysis. The participant pool was recruited through personal networks and social media and is skewed toward technically engaged users (64% Technology/IT/Engineering). The observed accessibility gains should not be generalized to novice users, low-AI-experience populations, or non-technical domains without further study. A larger study (N>200) with controlled task assignment and broader demographics would enable statistical testing of interaction effects.

**Model snapshot.** Results reflect model capabilities at the time of testing (March 2026). As models improve, the magnitude of structured encoding benefits may shift, though the qualitative pattern (weak-model compensation, language invariance) is likely to persist.

**User study order effect (fundamental design constraint).** The within-subject user study used a fixed sequence (A → D_raw → D_mod) without counterbalancing. This is a fundamental limitation rather than a minor confound: the sequence is monotonically increasing in intervention strength, participants perform the same self-selected task three times, and each successive round benefits from prior familiarity with both the task and the AI output. The observed improvement (3.16 → 3.66 → 4.04) almost certainly reflects a mixture of structural benefit, repeated-task clarification, and participant learning — and the current design cannot disentangle these. Future studies should employ counterbalanced or between-subject designs to establish causal attribution.

**User study vs. experiment model mismatch.** The controlled experiment (Section 4) uses international frontier models (Claude, GPT-4o, Gemini), while the user study (Section 6) reflects domestic Chinese AI usage (DeepSeek 44%, Doubao 32%). This mismatch means user study results cannot serve as direct ecological validation of the experimental findings on the same models. However, the consistency of structured prompting benefits across both model populations — frontier models in the experiment and domestic models in the user study — can be interpreted as evidence that the structured intent mechanism generalizes across model ecosystems.

**User study data quality.** A small number of participants misidentified the 5W3H expanded prompt as the final deliverable rather than as structured input to AI, introducing potential downward bias in D-condition ratings. One participant's task triggered AI safety refusal across all conditions, producing uniformly low scores unrelated to prompt structure. Neither observation alters the directional finding (A < D_raw < D_mod), which remained consistent across all sample sizes tested during data collection (N=10, 19, 46, 50).

**Ecological validity of framework comparison.** CO-STAR and RISEN prompts were authored by experimenters rather than practitioners of those frameworks, potentially underrepresenting their effectiveness in expert hands.

**Cross-paper Condition D comparability.** The AI expansion model for Condition D changed between studies: Papers 1-2 used DeepSeek-V3 (deepseek-chat API) while this study uses Qwen-Max (via lateni.com). This means Condition D prompts may differ in quality or style across papers, limiting the direct comparability of D-condition results across the series. Within this study, all D prompts are generated by the same pipeline, so internal comparisons remain valid.

## 8. Conclusion

This paper presents a large-scale empirical evaluation of structured intent encoding for human-AI interaction: 3,240 evaluated outputs across 3 models, 3 languages, 6 conditions, and 3 task domains, supplemented by a 50-person user study. To our knowledge, no existing evaluation provides this combination of cross-language coverage, multi-framework comparison, and intent-alignment scoring at this scale.

The central finding is that **structured intent encoding exhibits robust, largely language- and model-agnostic behavior** that reliably improves goal alignment across the conditions tested. Three independent frameworks (5W3H, CO-STAR, RISEN) achieve similarly high performance under the current evaluation resolution, suggesting that it is the act of dimensional decomposition — making implicit intent explicit — that drives alignment improvement, rather than any specific set of dimensions.

The weak-model compensation effect reveals that structured encoding is most valuable precisely where it is most needed: for weaker models and less experienced users. The 24× reduction in cross-language variance demonstrates that structured intent transcends linguistic surface form.

These findings motivate a shift in how we understand prompt engineering: not as a collection of techniques for individual tasks, but as the practical instantiation of an underlying intent transmission mechanism whose properties — dimensional decomposability, receiver compensation, and language invariance — are compatible with information-theoretic principles. As discussed in Section 7.6, the observed patterns are compatible with an information-theoretic interpretation: structured encoding can be viewed as decreasing the residual uncertainty of user intent given the prompt (conditional entropy reduction). Formalizing this interpretation into a predictive theoretical framework — one that can characterize optimal encoding strategies as a function of model capacity and task complexity — represents a natural next step for this line of research. These findings do not establish PPS as a communication protocol in the engineering sense, but they do support the view that structured intent representation may serve as a protocol-like communication layer for human-AI interaction.

Two limitations warrant mention when interpreting these results. First, the user study employed a fixed A → D_raw → D_mod sequence without counterbalancing; the observed satisfaction improvement partly reflects practice and learning effects alongside structural benefits. Second, the framework comparison (D≈E≈F) should be read as "all three saturate the current metric" rather than as confirmed true equivalence — finer-grained evaluation may yet reveal inter-framework differences. The consistency of structured prompting benefits across both frontier models (controlled experiment) and domestic Chinese AI models (user study) provides encouraging evidence that the mechanism generalizes across model ecosystems, but cross-ecosystem generalization should be tested more rigorously in future work.

All experimental data, evaluation scores, and analysis code are publicly available in the PPS-Bench dataset at github.com/PGlarry/prompt-protocol-specification.

**Conflict of Interest Statement.** The author is the creator of the PPS (Prompt Protocol Specification) framework and the lateni.com platform used to generate Condition D prompts in this study. As sole author, the author was responsible for all aspects of this work: framework design, experimental execution, data collection, analysis, and manuscript preparation, with no independent third-party involvement in any stage. The lateni.com platform is publicly accessible, and all generated prompts are included in the PPS-Bench dataset for independent verification and reproducibility. The experimental evaluation (goal alignment scoring) was performed by DeepSeek-V3, which is architecturally independent of the lateni.com expansion pipeline. The author has no financial relationship with any of the AI model providers (Anthropic, OpenAI, Google) whose models were evaluated in this study.

## About the Author

**PENG Gang** is the founder of Huizhou Lateni AI Technology Co., Ltd. and a researcher affiliated with Huizhou University, Huizhou, China. His research focuses on human-AI interaction, structured intent representation, and the design of communication protocols for AI systems. He developed the PPS (Prompt Protocol Specification) framework and the 5W3H-based intent encoding methodology, which form the foundation of the present series of empirical studies. He is also the creator of the lateni.com AI-assisted intent authoring platform, which is used to generate the AI-expanded prompts studied in this paper series.

Prior publications in this series include: "Evaluating 5W3H Structured Prompting for Intent Alignment in Human-AI Interaction" (arXiv:2603.18976, 2026) and "Does Structured Intent Representation Generalize? A Cross-Language, Cross-Model Empirical Study of 5W3H Prompting" (arXiv:2603.25379, 2026). His current research interests include the information-theoretic foundations of intent transmission in human-AI communication.

Contact: penggangjp@gmail.com | GitHub: github.com/PGlarry/prompt-protocol-specification